# PanicleNeRF: low-cost, high-precision in-field phenotyping of rice panicles with smartphone

**Authors**

Xin Yang[1,2], Xuqi Lu[1,2], Pengyao Xie[1,2], Ziyue Guo[1,2], Hui Fang[1], Haowei Fu[3], Xiaochun Hu[4], Zhenbiao Sun[4], Haiyan Cen[1,2*]

**Affiliations**

[1] College of Biosystems Engineering and Food Science, Zhejiang University, Hangzhou 310058, China
[2] Key Laboratory of Spectroscopy Sensing, Ministry of Agriculture and Rural Affairs, Hangzhou 310058, China
[3] Jiaxing Academy of Agricultural Science, Jiaxing, Zhejiang 314016, China
[4] Yuan Longping High-Tech Agriculture Co., Ltd, Changsha, 410128, China
* Correspondence: hycen@zju.edu.cn; Tel: +86 571 88982527.

**Abstract**

The rice panicle traits significantly influence grain yield, making them a primary target for rice phenotyping studies. However, most existing techniques are limited to controlled indoor environments and difficult to capture the rice panicle traits under natural growth conditions. Here, we developed PanicleNeRF, a novel method that enables high-precision and low-cost reconstruction of rice panicle three-dimensional (3D) models in the field using smartphone. The proposed method combined the large model Segment Anything Model (SAM) and the small model You Only Look Once version 8 (YOLOv8) to achieve high-precision segmentation of rice panicle images. The NeRF technique was then employed for 3D reconstruction using the images with 2D segmentation. Finally, the resulting point clouds are processed to successfully extract panicle traits. The results show that PanicleNeRF effectively addressed the 2D image segmentation task, achieving a mean F1 Score of 86.9% and a mean Intersection over Union (IoU) of 79.8%, with nearly double the boundary overlap (BO) performance compared to YOLOv8. As for point cloud quality, PanicleNeRF significantly outperformed traditional SfM-MVS (structure-from-motion and multi-view stereo) methods, such as COLMAP and Metashape. The panicle length was then accurately extracted with the rRMSE of 2.94% for *indica* and 1.75% for *japonica* rice. The panicle volume estimated from 3D point clouds strongly correlated with the grain number ($R^2$ = 0.85 for *indica* and 0.82 for *japonica*) and grain mass (0.80 for *indica* and 0.76 for *japonica*). This method provides a low-cost solution for high-throughput in-field phenotyping of rice panicles, accelerating the efficiency of rice breeding.

# MAIN TEXT

1. **Introduction**

Rice (Oryza sativa L.) is a crucial crop globally, feeding more than half of the world's population [1, 2]. Among the various factors influencing rice yield and quality, panicle traits such as panicle length, grain count, and grain mass are of great significance [3, 4]. Panicle traits are not only closely associated with rice yield but also serve as essential indicators in rice breeding [5, 6]. Therefore, approaches that enable high-precision measurements of rice panicle traits are crucial for accelerating rice breeding and improving overall crop productivity.

Numerous laboratory-based studies have been conducted on rice panicle phenotyping with machine vision technologies. Reported studies have employed different approaches, such as RGB scanning [7, 8], X-ray computed tomography [9, 10], structured-light projection [11], hyperspectral imaging [12], and multi-view imaging [13], to extract several panicle traits, including grain number, grain dimensions (length, width, and perimeter), kernel dimensions (length, width, and perimeter), seed setting rate, and panicle health status based on spectral signatures. While these methods are capable of phenotyping of rice panicle traits, they are limited to the controlled indoor environments and are still labor-intensive as panicles have to be harvested from the field and manually processed, including spreading the branches and fixing them in place. Thereby, development of high-throughput in-field phenotyping of rice panicles would be highly desired for improving efficiency and monitoring panicle development over time [14, 15].

In recent years, three-dimensional (3D) reconstruction methods have been increasingly employed for in-field plant phenotyping at different scales, offering more comprehensive and accurate information compared to 2D approaches. At the field scale, unmanned aerial vehicles (UAVs) equipped with RGB cameras [16] and airborne LiDAR [17] have been used to generate 3D point clouds of the entire field, which provide valuable insights into crop growth, canopy structure, and yield estimation. At the plot scale, terrestrial laser scanning (TLS) [18], depth cameras [19], and stereo vision systems [20] mounted on ground-based platforms have been utilized to reconstruct 3D models of plants within breeding plots , enabling the extraction of plant phenotypic traits and the assessment of genotypic differences. However, when it comes to the organ-level phenotyping, particularly for rice panicles, existing 3D reconstruction methods face significant challenges. The intricate structure and repetitive textures of rice panicles, combined with complex field conditions, present substantial challenges for accurate 3D reconstruction [21]. While handheld laser scanners can capture high-quality 3D data of individual panicles, the high cost and operational complexity with the required point clouds calibration limit their in-field application [22]. With the fast development of 3D reconstruction algorithms, it is hypothesized that multi-view imaging with a smartphone might provide a low-cost solution for 3D reconstruction of plants at organ-level, especially it would be accessible to almost everyone. To process the multi-view images captured by the smartphone, some 3D reconstruction methods can be employed. Among these, SfM-MVS, a traditional method combining structure-from-motion (SfM) [23] and multi-view stereo (MVS) [24], is a widely used 3D reconstruction technique. It has proven effective in generating 3D models in various applications with embedded into commercialized software such as COLMAP [25] and Metashape (Agisoft LLC, St. Petersburg, Russia). However, when applied to individual rice panicles, SfM-MVS often fails to generate complete and detailed point clouds due to the limitations of feature matching and dense reconstruction algorithms. Therefore, there is an urgent need for innovative 3D reconstruction of small plant organs such as rice panicles.

Neural radiance fields (NeRF) [26] as a novel 3D reconstruction technique has shown great potential for overcoming the limitations of traditional SfM-MVS methods for reconstructing high-quality 3D models [27–29]. NeRF represents a scene as a continuous function of 3D coordinates and viewing directions [30, 31], enabling the rendering of photorealistic novel views from a sparse set of input images [32, 33]. Recent advancements in NeRF, which include enhanced rendering quality and increased rendering speeds, suggest its potential for efficiently reconstructing detailed 3D models of rice panicles with less computational time [34–37]. However, NeRF mainly focuses on the scene reconstruction, and it lacks capabilities for object detection and segmentation. To accurately extract phenotypic traits of rice panicles, it is necessary to remove background elements and segment the target panicles from the reconstructed 3D scene. This requires the development of a high-quality image segmentation method that can be integrated with the NeRF model to enable precise detection and segmentation of rice panicles in the field.

The overall goal of this study is to develop a novel in-field phenotyping method by combining NeRF with You Only Look Once version 8 (YOLOv8) [38] and Segment Anything Model (SAM) [39], called PanicleNeRF, to extract rice panicle traits in the field accurately using a smartphone. The specific objectives are: (1) to accurately segment target rice panicles and label from 2D images, (2) to reconstruct complete, high-precision, and low-noise 3D models of rice panicles, (3) to preprocess and calibrate the point clouds, and (4) to extract panicle length and volume traits from the calibrated point cloud models and predict panicle grain number and grain mass.

## 2. Materials and Methods
### 2.1 Field experimental design and rice plant materials

Two field experiments were conducted during the ripening stage of rice crops at the breeding sites of Longping High-Tech in Lingshui, Hainan Province on May 20, 2023 (Exp 1) and Jiaxing Academy of Agricultural Sciences in Jiaxing, Zhejiang Province on November 15, 2023 (Exp 2). *Indica* rice (S616-2261-4/Hua Hui 8612) and *japonica* rice crops (Pigm/Zhejiang *japonica* 99) were planted at Exp 1 and Exp 2, respectively. The *indica* rice was transplanted with a spacing of 8 inches in length and 5 inches in width, with two seedlings per hole, while the *japonica* rice was spaced 6 inches in length and 5 inches in width, with one seedling per hole, as shown in Fig. 1 (a-b).

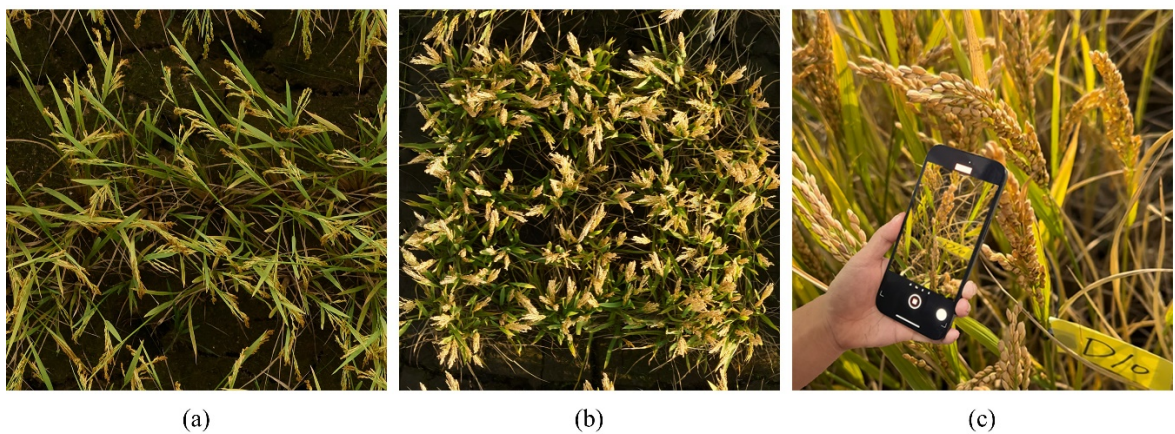

(a)　　　　　　　　　　　(b)　　　　　　　　　　　(c)

**Fig. 1.** Experimental fields and data acquisition. (a) Ripening stage of *indica* rice at the field of Longping High-Tech in Lingshui, Hainan Province, China. (b) Ripening stage of *japonica* rice at the field of Jiaxing Academy of Agricultural Sciences in Jiaxing, Zhejiang Province, China. (c) Data acquisition by circling around the target rice panicle using a smartphone.

## 2.2 Data acquisition

In-field data acquisition was conducted on 50 rice panicles at Exp 1 and Exp 2, respectively. The target rice panicles were randomly selected, and a label with known dimensions was affixed to each panicle for the size calibration. A 15 s video with the size of 1920×1080 pixels and a frame rate of 30 frames per second was then recorded by circling around the target rice panicle using a smartphone, as shown in Fig. 1 (c). Images were extracted from the collected video at a frequency of 15 frames per second to obtain multi-view images of each rice panicle. 225 original images were finally obtained for each rice panicle sample. After video recording, rice panicles were cut and brought back to the laboratory for specific trait measurements as the ground truth. A steel ruler was used to measure the panicle length, and the number of grains and grain mass per panicle were measured using a Thousand-Grain Weight Instrument developed by the DAAI team at Zhejiang University [40].

## 2.3 Data preprocessing

Data preprocessing was performed to compute the camera position and orientation as well as filter out images that did not adequately direct towards the target rice panicle to reduce the image processing time. SfM was employed to calculate the position and orientation of the camera. Pose-derived viewing direction (PDVD) and image alignment filtering (IAF) are proposed for image filtering. PDVD constructed a scene center point based on the camera's position and orientation. IAF calculated the angle between the line from the camera to the center point and the camera's orientation, filtering out images with an angle greater than 20°. For each sample, the number of images ranged from168 to 255 after filtering.

## 2.4 2D image segmentation

The 2D segmentation process was a crucial step in the development of PanicleNeRF. The purpose was to remove the background, enabling subsequent steps to generate point clouds including only the target rice panicle and the label. An ensemble model by integrating a large model with a small model was proposed in this study (Fig. 2. b). The ensemble model utilized the SAM for generic segmentation of the images. The segmentation results were then filtered based on area and stability, regions with an area smaller than 10,000 pixels or with a stability score below 0.8 were removed to eliminate the noise. To compensate for the lack of semantic information in the segmentation results by SAM, a trained YOLOv8 model was employed for instance segmentation, which generated rough masks of the target rice panicle and label. These rough masks were eroded, sampled, and matched with the fine masks from SAM to achieve precise instance segmentation. Finally, the segmented images were processed to remove backgrounds, generating transparent PNG images and completing the 2D segmentation.

# PanicleNeRF Method

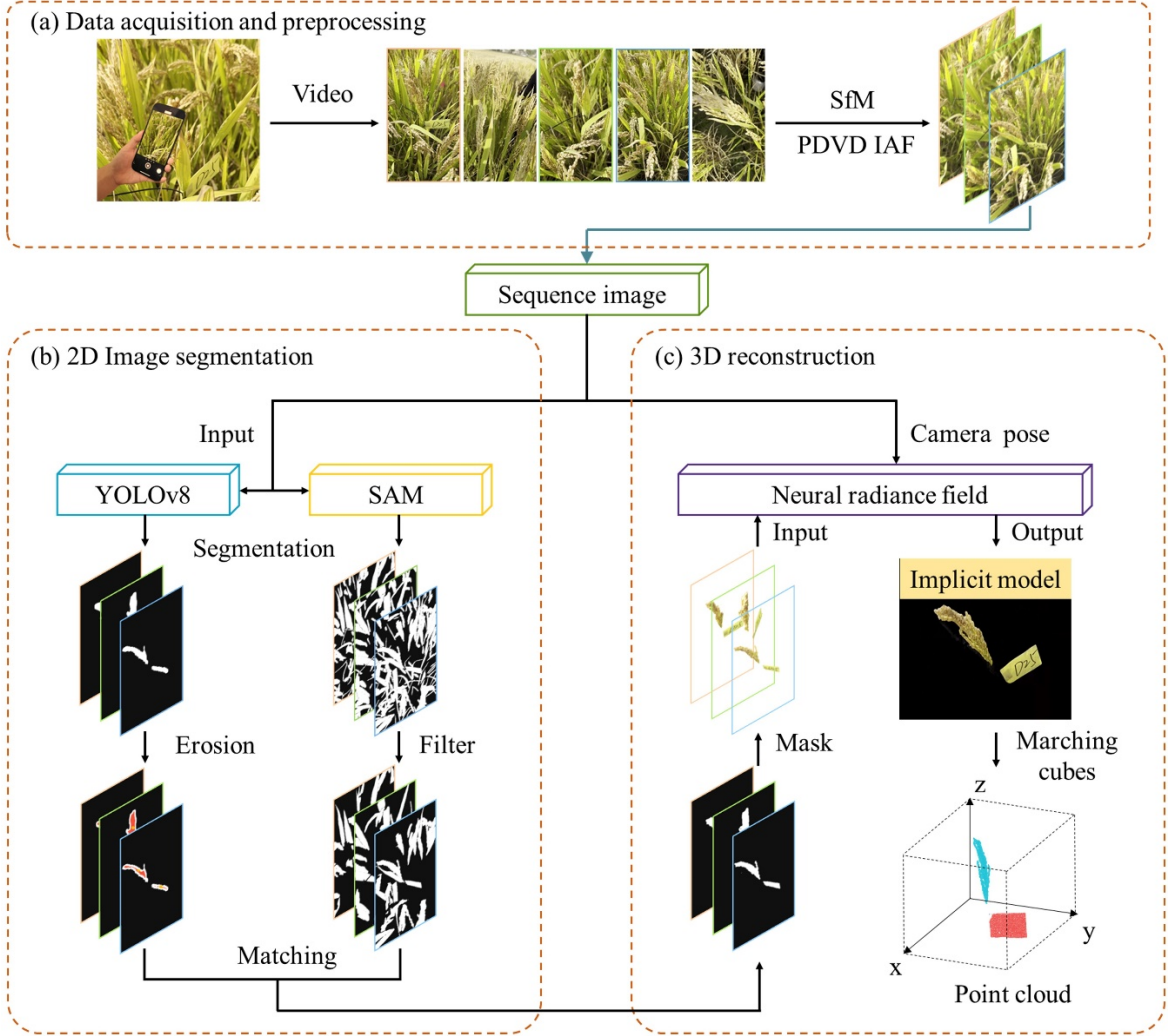

**Fig. 2.** Flowchart of PanicleNeRF method. (a) Data acquisition and preprocessing. (b) 2D image segmentation. (c) 3D reconstruction.

The performance of 2D image segmentation was evaluated with F1 score, Intersection over Union (IoU), and boundary overlap (BO), which are defined as follows:

$$Precision = \frac{TP}{TP+FP}$$

$$Recall = \frac{TP}{TP+FN}$$

$$F1 = \frac{2\,Precision \times Recall}{Preicision + Recall}$$

$$IoU = \frac{TP}{TP+FP+FN}$$

$$BO = \frac{|\,Ep \cap Eg\,|}{|\,Ep \cup Eg\,|}$$

where TP represents the number of true positive pixels, FP represents the number of false positive pixels, FN represents the number of false negative pixels, $Ep$ represents the set of predicted edge pixels, $Eg$ represents the set of ground truth edge pixels, ∩ denotes the intersection, and ∪ denotes the union. In addition, the PanicleNeRF method was also compared with well-known instance segmentation methods, YOLOv8 and mask region-based convolutional neural network (Mask-RCNN) [41].YOLOv8 is an object detection and instance segmentation method that utilizes a single-stage architecture for efficient and accurate predictions. It is an improved version of the YOLO series, offering enhanced performance and flexibility. While Mask-RCNN is a two-stage instance segmentation algorithm that extends the faster R-CNN framework by adding a branch for predicting segmentation masks. It has been widely used in both object detection and instance segmentation tasks by using a region proposal network (RPN) and a network head for mask prediction.

**2.5 3D reconstruction of rice panicles**

In this study, instant neural graphics primitive (Instant-NGP), a type of NeRF model, was employed for 3D reconstruction. It utilizes a multi-resolution hash encoding and spherical harmonics to enable fast and efficient train and inference [34]. The 3D reconstruction process is illustrated in Fig. 2 (c). The inputs comprised multi-view PNG images of rice panicles and their corresponding camera positions and orientations. The Instant-NGP algorithm was applied to learn a compact NeRF model representing the 3D structure of the target panicle and label. To streamline the data processing pipeline, the marching cubes algorithm was integrated with mesh sampling, allowing direct export of point clouds from the learned NeRF model [42]. The trained NeRF model was finally converted into a dense point cloud using this optimized marching cubes algorithm, resulting in point clouds of the rice panicle and label without background elements.

The 3D point clouds of rice panicle generated by PanicleNeRF was evaluated by the correlation with ground truth phenotypic traits. Specifically, the coefficient of determination (R2), root mean square error (RMSE) and relative root mean square error (rRMSE) between the predicted panicle length and the ground truth was calculated to validate the accuracy of the point cloud. R2 measures the proportion of variance in the ground truth explained by the predicted values, while RMSE and rRMSE quantify the average magnitude of prediction errors in absolute and relative terms, respectively. These metrics provide a comprehensive evaluation of the accuracy and reliability of the PanicleNeRF method for 3D reconstruction of rice panicles.

**2.6 Extraction of rice panicle traits**

2.6.1 Point cloud processing

Point cloud clustering was conducted using the density-based spatial clustering of applications with noise (DBSCAN) method [43]. It started with identifying core points by comparing the number of points within a predefined neighborhood to a set threshold. Then, points were assessed to determine if they were in the same cluster as a core point based on density-reachable relationships, effectively clustering the point cloud and removing insufficiently sized clusters deemed as outliers. After clustering, principal component analysis (PCA) was utilized to differentiate between the label and rice panicle clusters based on normal vector features, with the former marked in red and the latter in blue [44].

The size calibration of point clouds was performed based on a reference label with known dimensions. A bounding box algorithm was employed to locate the smallest enclosing rectangle of the point cloud representing the label, which had three faces with different areas. The point cloud was then projected onto the face with the median area and was fitted to calculate the length of the label in the point cloud scene. The size calibration was ultimately achieved by calculating the ratio between the length of the label and its corresponding length in the point cloud scene.

2.6.2 Extraction of panicle length trait

The point cloud of the target rice panicle was downsampled for rapid skeleton extraction and length calculation. As the standard Laplacian-based contraction (LBC) skeleton method was difficult to identify the start and end points of the rice panicle skeleton, a multi-tangent angle constraint was incorporated to avoid treating the tips of the rice panicle branches as endpoints [45]. After that, the skeleton was fitted with a curve to calculate the length of the panicle within the point cloud scene, and the panicle length trait was determined using the following formula:

$$L = L_1 \times \frac{X}{X_1}$$

where $L$ is the predicted length of the rice panicle, $L_1$ is the length of the rice panicle in the point cloud scene, $X$ is the length of the label, which equals 7.5 cm and $X_1$ is the length of the label in the point cloud scene.

2.6.3 Extraction of panicle volume trait

After the size calibration of point clouds, the rice panicle point cloud was voxelized with a voxel size of 0.01 units. The number of voxel units encompassed by the rice panicle point cloud was then calculated to determine the panicle volume using the following formula:

$$V = Num \times (0.01)^3 \times X^3$$

where $V$ is the predicted volume of the rice panicle, $Num$ is the number of voxel units encompassed by the rice panicle point and $X$ is the length of the label, which equals to 7.5 cm.

## 2.7 Development of a web-based platform

A user-friendly web-based platform (http://www.paniclenerf.com) was developed to enable researchers to upload videos of rice panicles recorded under windless or gentle wind conditions using smartphones or other RGB cameras. The videos should be captured by circling around the target rice panicle with a 7.5 cm label attached to the panicle for size calibration. The intuitive interface of the platform allows users to input their videos and retrieve the analyzed results, making the PanicleNeRF method accessible to a wider audience in the agricultural research community.

3. **Results**

**3.1 2D image segmentation**

Table 1 presents the results of three different 2D image segmentation methods for *indica* and *japonica* rice varieties. The proposed method, PanicleNeRF, outperformed the baseline method Mask-RCNN. For *indica* rice, PanicleNeRF achieved an F1 Score of 87.3% and 84.0% on the validation and test datasets, representing an improvement of 2.3% and 1.4%. Similarly, PanicleNeRF attained an IoU of 81.1% and 78.6%, surpassing Mask-RCNN by 4.8% and 5.1%. PanicleNeRF also demonstrated superior performance on *japonica* rice, with an F1 Score of 89.2% and 87.0%, exceeding Mask-RCNN by 3.7% and 1.2%. The IoU reached 81.3% and 78.2%, outperforming Mask-RCNN by 5.6% and 2.3%. Moreover, PanicleNeRF performed slightly better than YOLOv8 across all metrics for both rice varieties.

**Table 1.** The comparison of 2D image segmentation methods for *indica* and *japonica* rice varieties. The best results are in boldface.

| Variety | Method | F1 Score (%) | | IoU (%) | |
|---|---|---|---|---|---|
| | | Val | Test | Val | Test |
| Indica rice | Mask-RCNN | 85.0 | 82.6 | 76.3 | 73.5 |
| | YOLOv8 | 86.1 | 82.9 | 79.2 | 76.4 |
| | PanicleNeRF | **87.3** | **84.0** | **81.1** | **78.6** |
| Japonica rice | Mask-RCNN | 85.5 | 85.8 | 75.7 | 75.9 |
| | YOLOv8 | 88.7 | 85.3 | 80.9 | 77.4 |
| | PanicleNeRF | **89.2** | **87.0** | **81.3** | **78.2** |

The boundary segmentation performance is presented in Figure 3. PanicleNeRF achieved the best BO results in *indica* with 8.5% for both Validation and Test, and in *japonica* with 9.5% for Validation and 10.0% for Test. These results showed improvements of 4.0%, 3.9%, 5.1% and 5.6% over the second-best results obtained by YOLOv8. Notably, the interquartile range (IQR) of PanicleNeRF's performance was entirely above that of Mask-RCNN and YOLOv8 across all four categories. Figure 4 illustrates the representative results of the three methods on the testing set. Despite the complex background and branching structure of rice panicles, PanicleNeRF achieved excellent segmentation performance, while Mask-RCNN displayed significant fluctuations in edge segmentation and YOLOv8 produced noticeable jagged edges when segmenting the branches. The substantial improvement in the BO value distribution, along with the enhanced segmentation results, demonstrated the superior performance and robustness of the PanicleNeRF method.

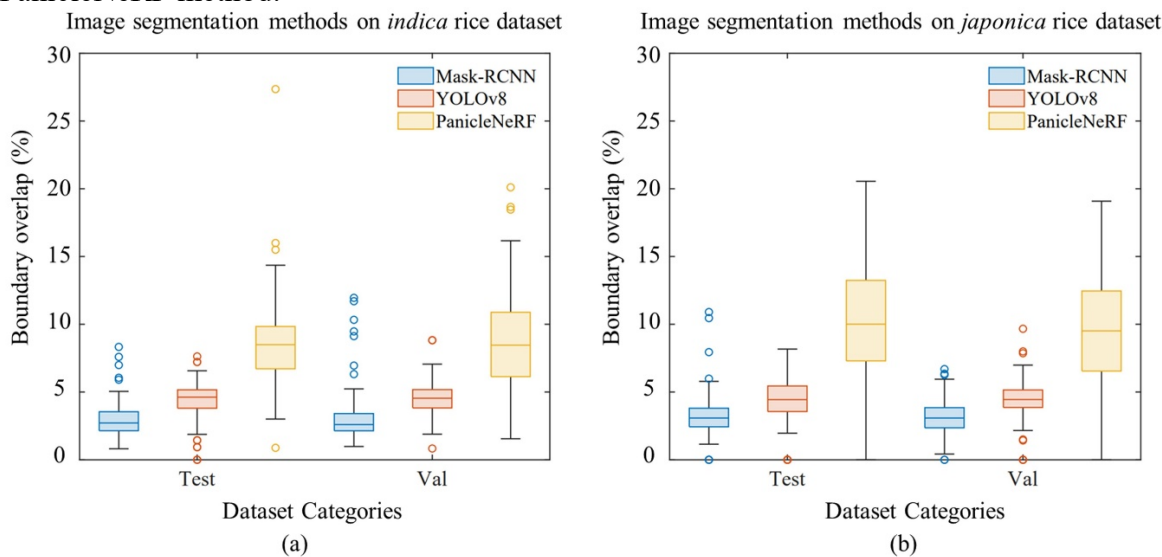

**Fig. 3.** The boundary overlap performance of different methods on rice varieties. (a) Performance on *indica* rice dataset. (b) Performance on *japonica* rice dataset.

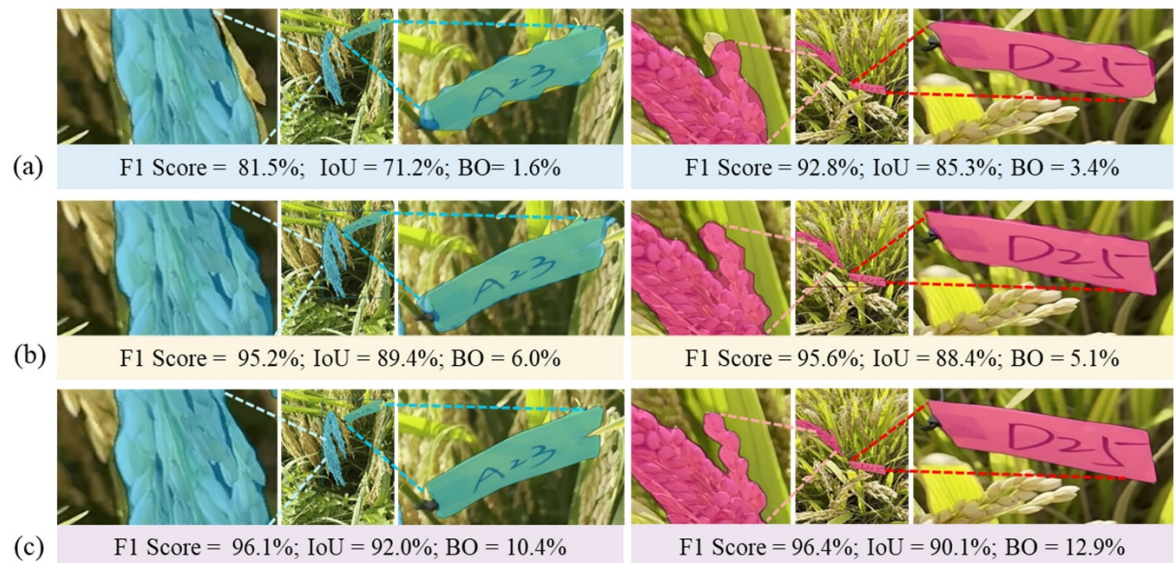

**Fig. 4.** Illustration of the representative image segmentation results on *indica* rice (left column) and *japonica* rice (right column) by different methods. (a) Mask-RCNN segmentation results. (b) YOLOv8 segmentation results. (c) PanicleNeRF segmentation results.

## 3.2 3D reconstruction of rice panicle

Fig. 5 presents the background removal results in the NeRF model after using the PanicleNeRF method for representative samples of *indica* and *japonica* rice. (a) and (c) show the NeRF model obtained by inputting the original image set, while (b) and (d) display the NeRF model obtained by inputting the same set of images after 2D segmentation. The results demonstrate that PanicleNeRF effectively removed non-regions of interest and extracts the target rice panicles and labels in the NeRF model. Furthermore, the clustering results of the panicle and label point clouds are illustrated in Fig. 6, where (a) and (b) represent *indica* and *japonica* rice. In both cases, the blue points indicate the panicle semantics, while the red points indicate the label semantics. The clear separation of the panicle and label point clouds demonstrate the effectiveness of the PanicleNeRF in distinguishing between the two semantic categories. This effective clustering lays the foundation for subsequent extraction of rice panicle traits.

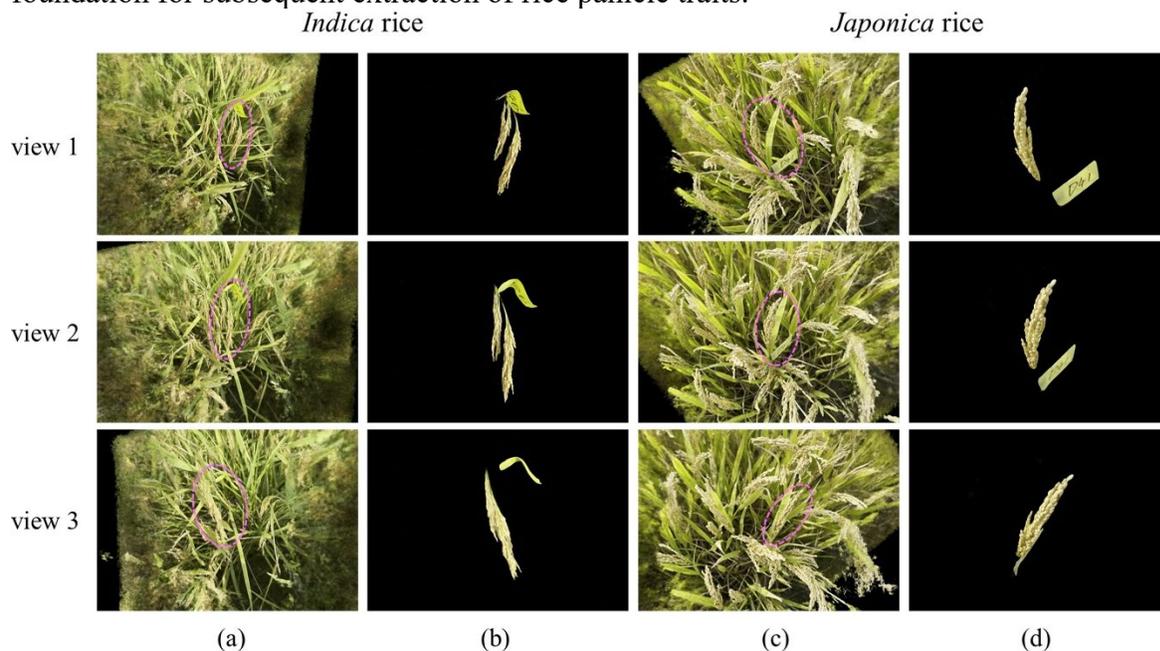

**Fig. 5.** NeRF models of *indica* and *japonica* rice reconstructed using original image set and 2D segmented images, viewed from three different perspectives of the NeRF model. (a) *Indica* NeRF model obtained from the original image set. (b) *Indica* NeRF model obtained from images with 2D segmentation. (c) *Japonica* NeRF model obtained from the original image set. (d) *Japonica* NeRF model obtained from images with 2D segmentation.

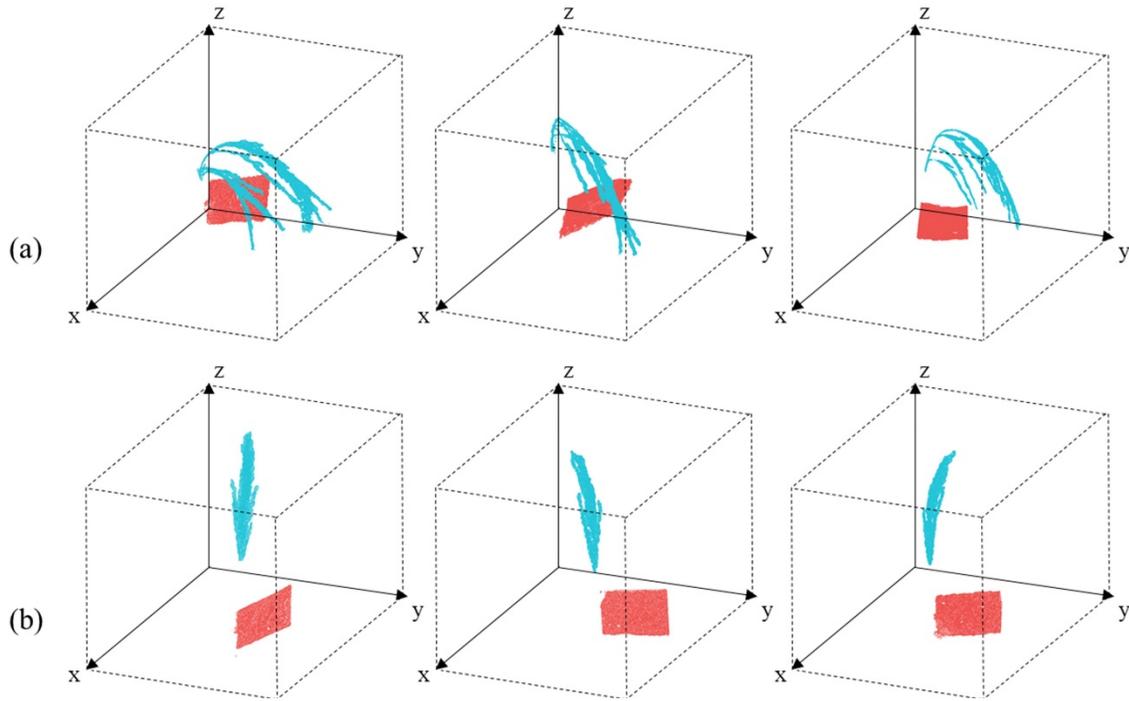

**Fig. 6.** Illustration of the representative clustering results of rice panicle and label point clouds for (a) *indica* rice and (b) *japonica* rice, where blue indicates the panicle semantics and red indicates the label semantics.

Fig. 7 presents a comparison of the point clouds generated by PanicleNeRF with those generated by COLMAP and Metashape. (a) and (b) show the front and side views of the *indica* rice point cloud, while (c) and (d) display the front and side views of the *japonica* rice point cloud. Overall, the point cloud generated by PanicleNeRF was the most complete, with the fewest point cloud holes and the highest resolution. The side view of the *indica* rice point cloud (b) clearly illustrates that the point cloud generated by COLMAP had more noise and could not distinguish the branches of the rice panicle, while the point cloud generated by Metashape was more fragmented and had poor completeness. The other viewpoints also demonstrate the fragmentation and incompleteness of COLMAP and Metashape point clouds in the reconstruction of rice panicles and labels, which leads to difficulties in their practical application for field phenotype extraction.

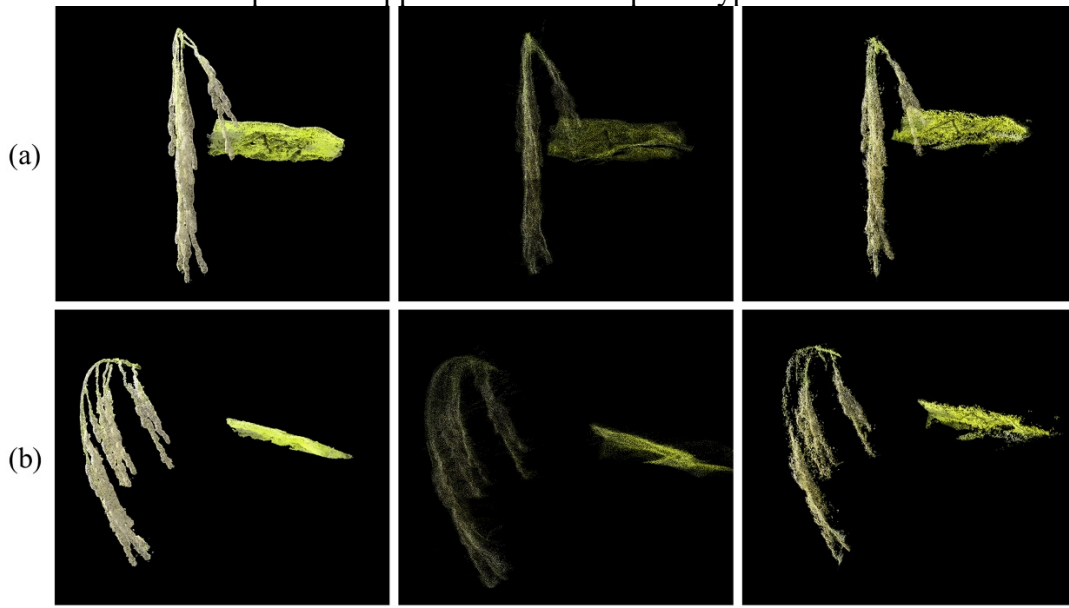

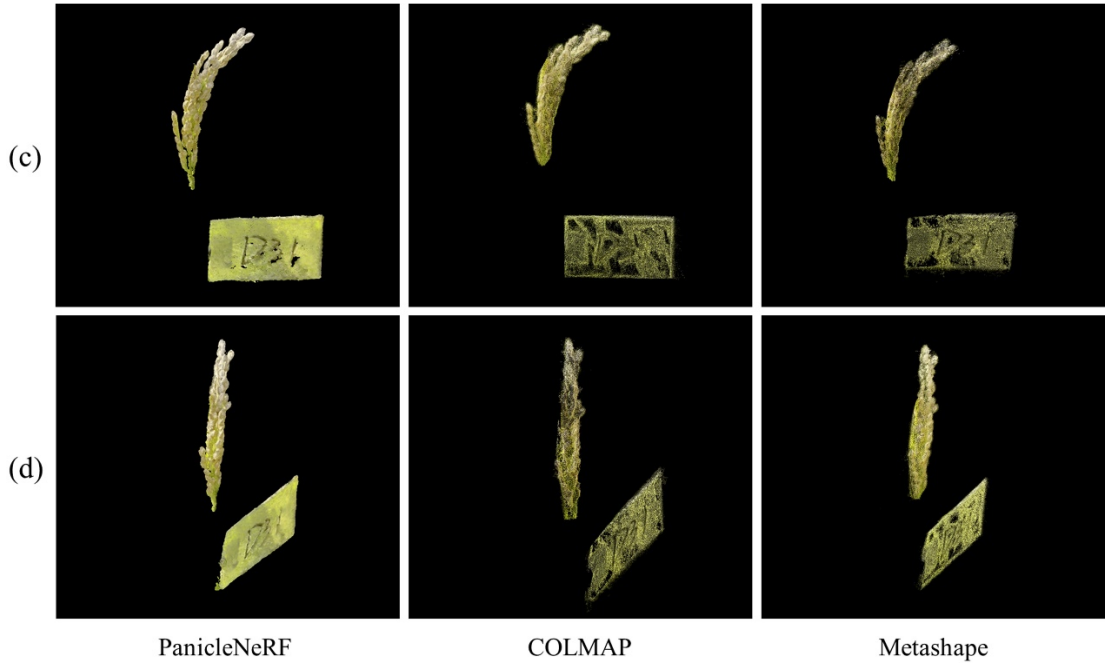

|     | PanicleNeRF | COLMAP | Metashape |

**Fig. 7.** The comparison of representative rice panicle point clouds reconstructed by PanicleNeRF and traditional 3D reconstruction methods (COLMAP and Metashape). (a) Front view of *indica* rice. (b) Side view of *indica* rice. (c) Front view of *japonica* rice. (d) Side view of *japonica* rice.

### 3.3 Panicle traits extraction

Fig. 8 shows the correlation between the predicted and measured panicle lengths for *indica* and *japonica* rice. For *indica* rice, the $R^2$ value was 0.90, the RMSE was 0.71, and the rRMSE was 2.94%, while for *japonica* rice, the $R^2$ value was 0.95, the RMSE was 0.26, and the rRMSE was 1.75%. These results validated the effectiveness of the size calibration and the accuracy of the 3D point clouds. The results also indicated that the prediction accuracy for *japonica* rice was higher than that for *indica* rice, as the $R^2$ value was higher and the rRMSE was lower for *japonica* rice.

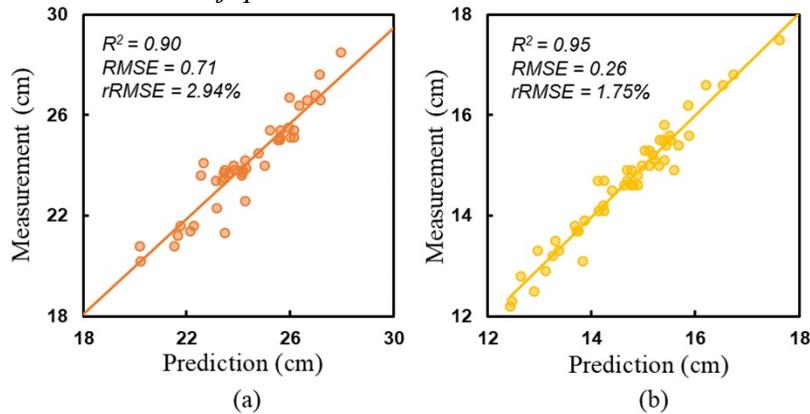

**Fig. 8.** Correlation analysis between predicted and measured panicle lengths for (a) *indica* rice and (b) *japonica* rice.

Fig. 9 shows the correlation between the predicted panicle volume and the measured grain count and grain mass for *indica* and *japonica* rice. For *indica* rice, the $R^2$, RMSE and rRMSE values were 0.85, 15.61 and 7.45% for grain count, and 0.80, 0.45 and 8.51% for grain mass. For *japonica* rice, the corresponding values were 0.82, 15.42 and 9.74% for grain count, and 0.76, 0.47 and 10.33% for grain mass. These results demonstrated a strong correlation between the predicted panicle volume and grain count, as well as grain mass. Besides, the prediction accuracy for both grain count and grain mass was higher in *indica* rice compared to *japonica* rice, with *indica* rice having higher $R^2$ values and lower rRMSE values for both traits.

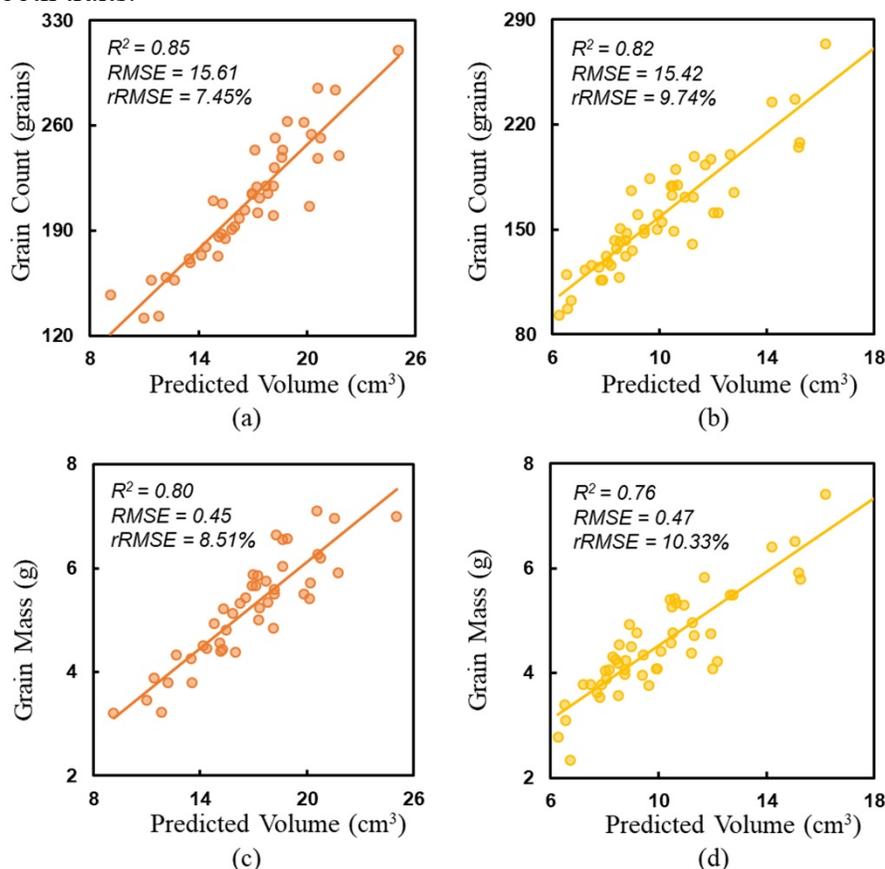

**Fig. 9.** Correlation analysis between predicted panicle volume and measured grain count and grain mass for *indica* and *japonica* rice. (a) Predicted volume versus measured grain count for *indica* rice. (b) Predicted volume versus measured grain count for *japonica* rice. (c) Predicted volume versus measured grain mass for *indica* rice. (d) Predicted volume versus measured grain mass for *japonica* rice.

4. **Discussion**
 **4.1 Multi-model fusion outperforms single models in rice panicle 2D segmentation**
   The results demonstrated that the proposed 2D rice panicle segmentation method, which combined the advantages of the large model SAM and the small model YOLOv8, achieved superior performance compared to using a single model. As illustrated in Figure 4, utilizing a single model for segmentation often led to incomplete target contours, while the accuracy of the 3D reconstruction method employed in this research relied on the segmentation quality of 2D images. Existing models struggled to achieve satisfactory results because they were primarily designed to perform well on public datasets such as ImageNet, where crop categories accounted for only around 2.4% of the total image categories, with the majority of targets being common rigid objects encountered in daily life and exhibiting distinct features compared to crops [46]. It is unsurprising that using transfer learning to fine-tune existing models on crop datasets caused suboptimal segmentation contours [47]. However, training a new network specifically tailored for crop segmentation requires a substantial amount of samples, and currently, there is a lack of sufficient data to train a model capable of achieving high-precision contour segmentation.

   With the advancement of ultra-large models such as SAM, their generalization ability has significantly improved, enabling accurate contour segmentation even for targets not included in the training set. Nevertheless, these models could not perform individual segmentation for specific targets without appropriate prompts. It is precisely by leveraging the strengths of both large and small models that our proposed method achieved superior segmentation results. Targets with inaccurate segmentation can be used as prompts and fed into large models, thereby obtaining targets with precise contour segmentation.

### 4.2 Superiority of PanicleNeRF in fine-scale

Traditional SfM-MVS-based methods, such as COLMAP and Metashape, produce noisy and incomplete point clouds when dealing with in-field rice panicle scenes. This suboptimal performance could be attributed to the following reasons. Firstly, objects with repetitive textures, such as rice panicles, are prone to feature matching errors due to the high similarity of feature points at different positions, leading to incorrect matching and large deviations in the recovered 3D point positions [48]. Secondly, Small objects present challenges for accurate 3D structure recovery due to inadequate parallax information. The small disparities of small objects across different views lead to large depth estimation errors in triangulation and 3D point recovery. Lastly, SfM-MVS relies on geometric constraints between images, which are often not satisfied when objects have occlusions and complex shapes [49]. Complex shapes reduce the correspondence between images, while occlusions obstruct feature matching, resulting in a lack of effective geometric constraints and compromising the completeness and precision of the reconstructed objects.

In contrast, PanicleNeRF achieved the least noisy and most complete point cloud compared to the SfM-MVS-based method. The superiority of PanicleNeRF can be attributed to its ability to learn the geometric structure and appearance information of the scene through neural networks. Unlike traditional methods that rely on feature point matching and triangulation, PanicleNeRF directly learns the 3D structure from images, enabling it to handle situations with similar features, repetitive textures, and insufficient parallax. Furthermore, by representing the scene through continuous density and color fields, PanicleNeRF can generate more complete and detail-rich reconstruction results. The significant advantages demonstrated by PanicleNeRF in fine-grained 3D reconstruction tasks under complex scenes highlight its potential to advance plant phenotypic analysis.

### 4.3 Panicle traits extraction performance and comparisons

The results demonstrate that high R2 and low rRMSE were achieved in panicle length, grain count and grain mass prediction. During this process, we discovered that the panicle length prediction for *indica* rice was not as accurate as that for *japonica* rice, possibly due to the more compact and convergent branches of *japonica* rice compared to the looser and more dispersed branches of *indica* rice. When extracting the skeleton, the central skeleton extraction of multi-branched panicles is more challenging than that of panicles with branches clustered together. Additionally, we found that when using panicle volume to predict grain count and grain mass, the *indica* rice dataset performed better than the *japonica* rice dataset. This is primarily attributed to the compact branches of *japonica* rice, which might have created gaps within the panicle that were treated as solid matter, potentially limiting the accuracy of the model. Such limitations are common in visible image-based reconstructions and may only be addressed by adopting other techniques like computed tomography or magnetic resonance imaging [50, 51].

In Fig. 10, the correlations between panicle length, panicle volume, grain count, and grain mass in the *indica* and *japonica* rice datasets were calculated and visualized. For *indica* rice, the correlations between panicle length and the other three phenotypes were only 0.37, 0.32, and 0.43, while the correlations between panicle volume and grain count and grain mass were 0.85 and 0.80, respectively. A similar pattern was observed for *japonica* rice. Panicle volume, which can be extracted by 3D reconstruction methods, showed much stronger correlations with grain count and grain mass compared to panicle length. This underscores the importance of accurate 3D models and their unique capability to extract panicle volume, providing valuable information that cannot be achievable with 2D models.

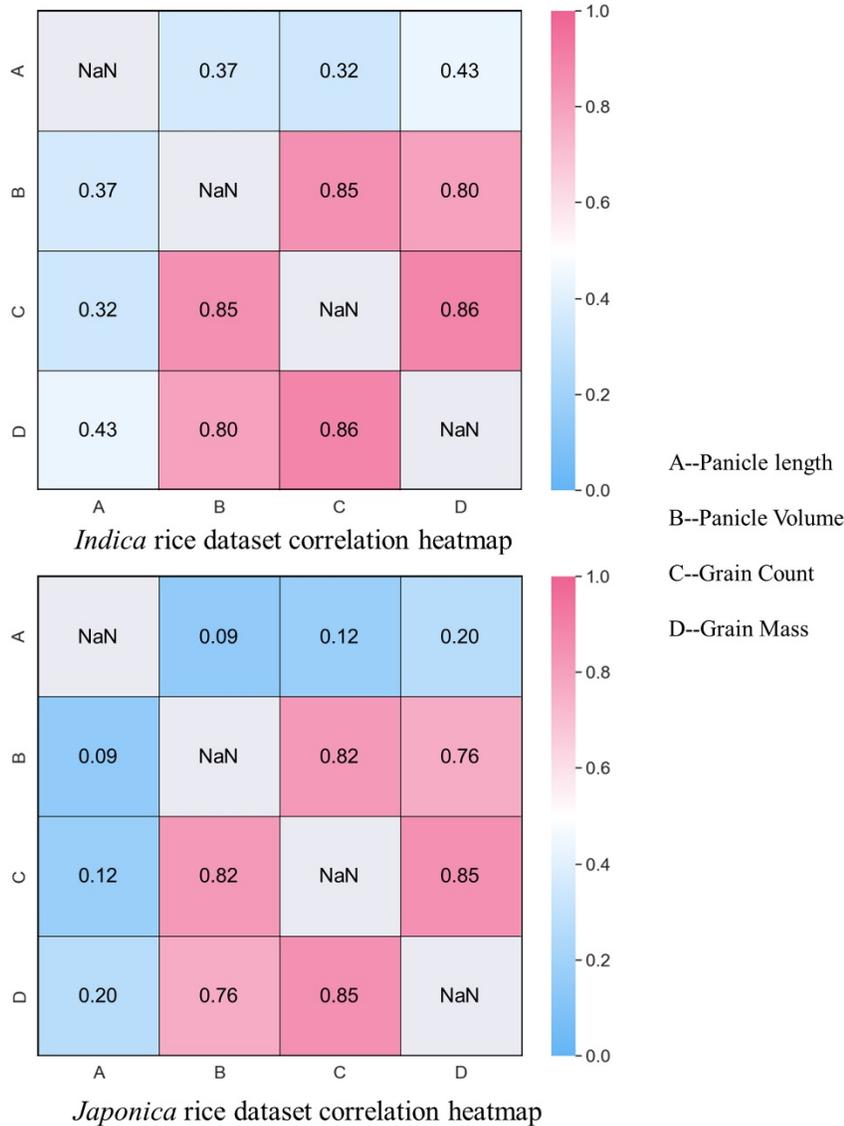

Fig. 10. Heatmap of correlations among panicle length, panicle volume, grain count, and grain mass for different rice panicle types in the *indica* and *japonica* datasets.

We also compared the results obtained in this study with those reported in previous indoor rice panicle phenotyping research. These methods have yielded R2 values ranging from 0.70 to 0.76 for grain count and mass prediction [52, 53]. In comparison, our PanicleNeRF method achieves better results, with R2 values of 0.83 for grain count and 0.78 for grain mass, demonstrating its effectiveness in field-based rice panicle phenotyping.

### 4.4 Limitation and future prospects

PanicleNeRF had achieved promising performance in rice panicle phenotyping. However, there were still some limitations of our method. One limitation was that the throughput was limited as it focused on extracting detailed 3D models of individual rice panicle. In the future, by extending PanicleNeRF, low-cost remote sensing equipment, such as UAVs, could potentially be employed to analyze high-throughput and large-scale fine-grained 3D models of rice panicles in the field [54–56]. In addition, the method required data acquisition under windless or light breeze conditions, which restricted its convenience of application in actual paddy fields. To address this issue, multiple cameras could be utilized for synchronous exposure, enabling simultaneous acquisition of multi-view high-quality imaging of rice plants, effectively reducing the interference of natural wind.

Moreover, the camera extrinsic parameter inference in data preprocessing was time-consuming, taking up to 15 minutes (Table 2). This could be improved in the future by designing a specialized data acquisition device that enables consistent relative positions and orientations of cameras in each acquisition, thereby utilizing fixed camera extrinsic parameters to avoid this time-consuming step.

Table 2 Time consumption analysis for each step in the PanicleNeRF workflow.

| Workflow step | Time consumed |
|---|---|
| Data acquisition | 15 s |
| Data preprocessing | 15 min |
| 2D image segmentation | 2 min |
| 3D reconstruction | 2 min |
| Extraction of rice panicle traits | 4 min |

5. **Conclusions**

The conventional 3D reconstruction and segmentation methods often generate noisy and fragmented point clouds when dealing with the complex structure and repetitive texture of rice panicles, which is not suitable for phenotyping panicles in the field. To address this challenging problem, we propose PanicleNeRF, a novel method that enables high-precision and low-cost reconstruction of rice panicle 3D models in the field using smartphone. The proposed method combined the large model SAM and the small model YOLOv8 to achieve high-precision segmentation of rice panicle images. The NeRF technique was then employed for 3D reconstruction using the images with 2D segmentation. Finally, the resulting point clouds are processed to successfully extract and analyze panicle phenotypes. The results show that PanicleNeRF effectively addressed the task of 2D image segmentation, achieving a mean F1 Score of 86.9% and a mean IoU of 79.8%, with nearly double the BO performance compared to YOLOv8. In terms of point cloud quality, PanicleNeRF significantly outperformed traditional SfM-MVS methods, such as COLMAP and Metashape. The panicle length was then accurately extracted with the rRMSE of 2.94% for *indica* and 1.75% for *japonica* rice. The panicle volume estimated from high quality 3D point clouds strongly correlated with the actual grain number ($R^2 = 0.85$ for *indica* and $R^2 = 0.82$ for *japonica*) and grain mass ($R^2 = 0.80$ for *indica* and $R^2 = 0.76$ for *japonica*). This work is expected to contribute to the advancement of high-quality in-field rice panicle phenotyping, facilitating the progress of rice phenotyping and breeding efforts.

**Acknowledgments**

**Author contributions:** X.Y., X.L., P.X. and Z.G. designed the study, performed the experiment, and wrote the manuscript. X.Y. developed the algorithm and analyzed the data. H.F. developed a web platform for the algorithms. H.F., X.H. and Z.S. cultivated two different rice varieties. H.C. supervised experiments at all stages and revised the manuscript.

**Funding:** This work was supported by the Fundamental Research Funds for the Central Universities (226-2022-00217), Key R&D Program of Zhejiang Province (2021C02057), and Zhejiang University Global Partnership Fund (188170+194452208/005).

**Competing interests:** The authors declare that there is no conflict of interest regarding the publication of this article.

**Data Availability:** The data is freely available upon reasonable request.